\documentclass[10pt,twocolumn]{article} 
\usepackage{simpleConference}
\usepackage{times}
\usepackage{graphicx}
\usepackage{amssymb}
\usepackage{url,hyperref}
\usepackage{booktabs}
\usepackage{float}

\graphicspath{{media/}}     

\begin{document}

\title{Engineering Artificial Intelligence: Framework, Challenges, and Future Direction}

\author{Jay Lee, Hanqi Su\thanks{\textit{Corresponding author}}, Dai-Yan Ji, and Takanobu Minami \\
Center for Industrial Artificial Intelligence, Department of Mechanical Engineering, \\
A. James Clark School of Engineering, University of Maryland, College Park, \\
Maryland, United States of America \\
\texttt{hanqisu@umd.edu}  \\
}

\maketitle
\thispagestyle{empty}

\begin{abstract}
Over the past ten years, the application of artificial intelligence (AI) and machine learning (ML) in engineering domains has gained significant popularity, showcasing their potential in data-driven contexts. However, the complexity and diversity of engineering problems often require the development of domain-specific AI approaches, which are frequently hindered by a lack of systematic methodologies, scalability, and robustness during the development process. To address this gap, this paper introduces the "ABCDE" as the key elements of Engineering AI and proposes a unified, systematic engineering AI ecosystem framework, including eight essential layers,  along with attributes, goals, and applications, to guide the development and deployment of AI solutions for specific engineering needs. Additionally, key challenges are examined, and eight future research directions are highlighted. By providing a comprehensive perspective, this paper aims to advance the strategic implementation of AI, fostering the development of next-generation engineering AI solutions.
\end{abstract}

\section{Introduction}
In the era of big data, artificial intelligence (AI) and machine learning (ML) solutions are increasingly being applied across diverse engineering fields, such as manufacturing~\cite{dogan2021machine,sahoo2022smart}, predictive maintenance~\cite{carvalho2019systematic, su2024machine}, energy~\cite{yao2023machine, entezari2023artificial}, transportation~\cite{veres2019deep, yuan2022machine}, healthcare~\cite{zhang2022shifting, singhal2023large}, design~\cite{wang2022machine, yuksel2023review}, computer vision (CV)~\cite{chai2021deep, guo2022attention}, natural language processing (NLP)~\cite{min2023recent, khurana2023natural}, construction~\cite{xu2021computer, baduge2022artificial}, robotics and automation~\cite{ribeiro2021robotic, soori2023artificial}, agriculture~\cite{sharma2020machine, benos2021machine}, materials science~\cite{batra2021emerging, cheetham2024artificial}, etc. The application of AI and ML in engineering domains is expected to revolutionize traditional practices, accelerating smarter, more efficient, and greener solutions~\cite{jan2023artificial,nti2022applications,lee2025rethinking}. However, the complexity and diversity of engineering problems necessitate the development of customized AI solutions to address specific challenges. Recognizing this gap, \textbf{Engineering AI} has emerged as a systematic discipline, bridging the gap between AI, ML, and engineering requirements. In this paper, key elements in engineering AI ("ABCDE") are identified. Beyond these elements, a structured ecosystem framework comprising eight essential layers is proposed to guide practitioners and researchers in developing robust and powerful engineering AI ecosystems. In addition, key challenges and future research directions are also discussed. Through this exploration, this perspective paper aims to provide strategic guidance for the use of AI in different engineering fields to promote innovation and practical applications of AI in engineering.

\begin{figure*}[!ht]
\centering\includegraphics[width=0.8\linewidth]{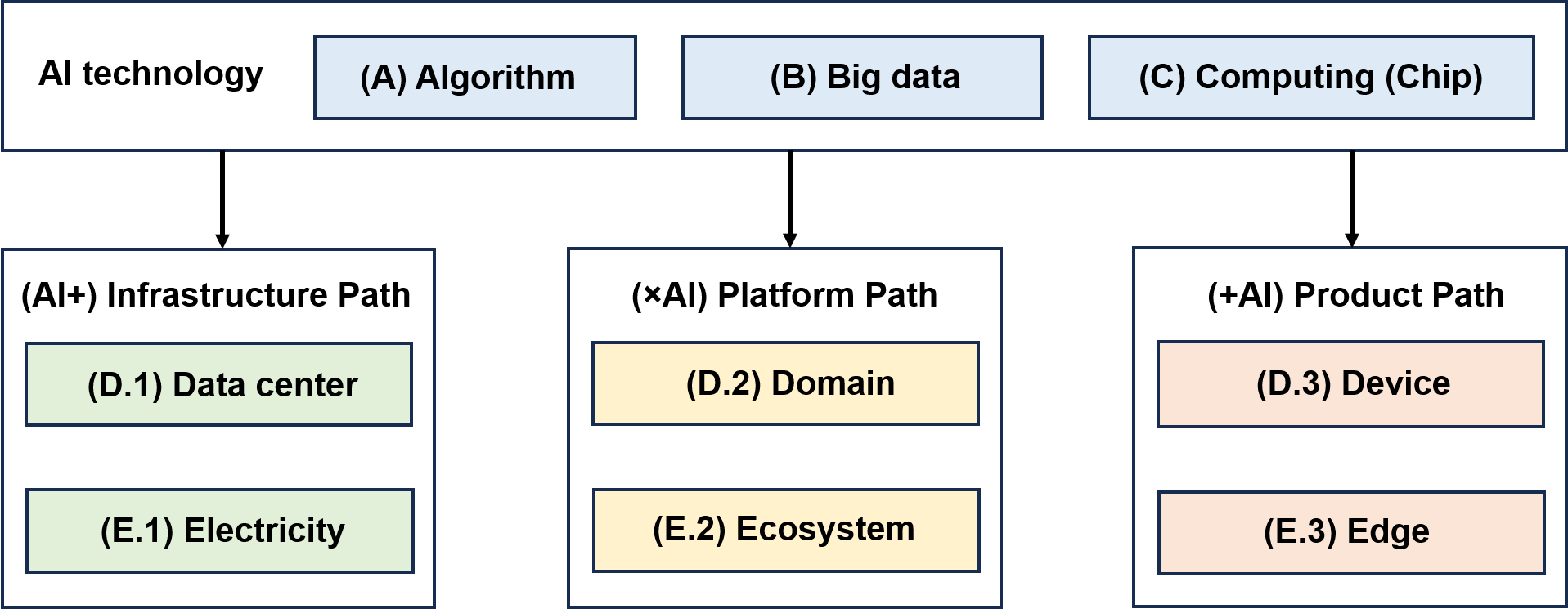}
\caption{Key Elements in Engineering AI: ABCDE.}
\label{fig:ABCDE}
\end{figure*}

\section{Key Elements in Engineering AI: ABCDE}
The foundational components of engineering AI can be categorized into the "ABCDE" as shown in Figure~\ref{fig:ABCDE}, where the first three components—Algorithm, Big Data, and Computing Power—form the key elements of AI technology. \emph{(A. Algorithm)} represents advanced AI and ML methods, \emph{(B. Big Data)} involves collection, storage, and processing of engineering-specific data, and \emph{(C. Computing)} refers to state-of-the-art computational infrastructure, including high-performance chips and cloud resources. Together, these elements build the foundation for AI-driven tasks in engineering. Beyond these "ABC" components, the "DE" pairs extend into three key paths: \emph{(AI+) Infrastructure, ($\times$AI) Platform, and (+AI) Product}. 1) (AI+) Infrastructure path emphasizes physical and energy systems, with \emph{(D.1 Data Center)} supporting AI computation and data storage, and \emph{(E.1 Electricity)} powering AI infrastructure. 2) ($\times$AI) Platform path focuses on industrial domain knowledge \emph{(D.2 Domain)} and AI-enabled ecosystems \emph{(E.2 Ecosystem)} development to create AI solutions for different engineering domains. 3) the (+AI) Product path bridges AI-powered devices \emph{(D.3 Device)} with edge computing environments \emph{(E.3 Edge)} to enable real-time processing and decision-making near the data source. While the ABC components form the core technological foundation required to develop AI systems, they influence each other dynamically. Advances in algorithms require greater computing power and more diverse data, while computational innovation and large amounts of domain-specific data drive the development of new algorithms. The DE components define where and how AI technologies enabled by ABC components are deployed and utilized. These paths define the practical landscape for engineering AI, where each path contributes a distinct role: enabling computation and energy support (Data Center, Electricity), integrating domain knowledge and ecosystems (Domain, Ecosystem), and deploying intelligent, embedded applications at the edge (Device, Edge).


\section{Conceptual Framework for Engineering Artificial Intelligence Ecosystem}
\begin{figure*}[!ht]
\centering\includegraphics[width=1.0\linewidth]{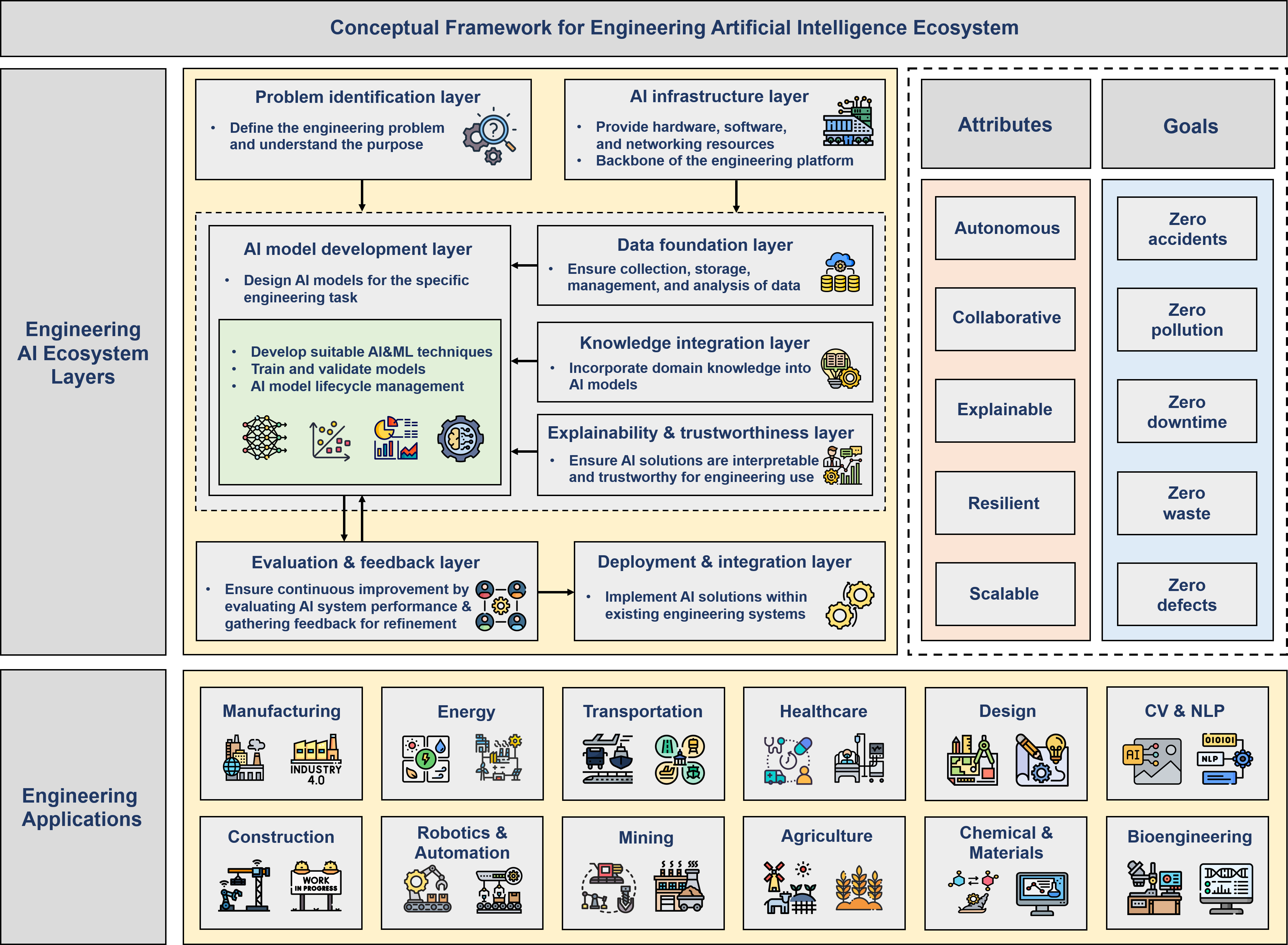}
\caption{Conceptual Framework for Engineering Artificial Intelligence Ecosystem.}
\label{fig:framework}
\end{figure*}

Figure~\ref{fig:framework} shows the proposed Engineering AI Ecosystem, which serves as a comprehensive guideline for designing, deploying, and refining AI solutions across different engineering domains. This framework provides practitioners with a systematic approach to strategizing the development and implementation of engineering AI systems. It comprises eight essential layers: \emph{$"$Problem identification layer$"$, $"$AI infrastructure layer$"$, $"$AI model development layer$"$, $"$Data foundation layer$"$, $"$Knowledge integration layer$"$, $"$Explainability \& trustworthiness layer$"$, $"$Evaluation \& feedback layer$"$, and $"$Deployment \& integration layers$"$.} These layers collectively ensure that future engineering AI systems are autonomous, collaborative, explainable, resilient, and scalable (\textbf{key attributes}), ultimately achieving zero accidents, zero downtime, zero waste, zero pollution, and zero defects (\textbf{engineering goals}). A detailed explanation of each layer is provided below.

\subsection{Problem identification layer}
The problem identification layer is the first step in the engineering AI ecosystem. This layer concentrates on defining the specific engineering problem that the AI system aims to address. It involves: (1) understanding the purpose and scope of the problem; (2) identifying technical constraints, pain points, potential challenges, and desired outcomes; (3) collaborating and conducting a detailed analysis to get insights. Ultimately, practitioners will be confident in using AI and ML techniques to solve the problem through subsequent layers.

\subsection{AI infrastructure layer}
The AI infrastructure layer, acting as the backbone of the conceptual framework, provides necessary hardware, software, and networking resources to support AI development and deployment. On the hardware side, this includes compute resources such as CPUs, GPUs, TPUs, cloud computing services, servers, and IoT devices. It also involves storage solutions, including high-speed storage, distributed file systems, and databases for efficient data storage and management. On the software side, the infrastructure supports various AI frameworks (e.g., TensorFlow, PyTorch), libraries (e.g., Scikit-learn), and development tools (e.g., Jupyter, Google Colab) to facilitate data processing and model development. Additionally, networking resources ensure efficient communication between components and seamless data exchange across systems. Practitioners should select appropriate platforms and resources based on their specific problem requirements to optimize cost, time, and efficiency. 

\subsection{Data foundation layer}
The data foundation layer, a necessary prerequisite for the "AI model development layer", involves the entire process of data collection, storage, management, and analysis, resulting in high-quality datasets suitable for AI applications~\cite{whang2023data,li2024data}. After data acquisition, which is supported by the "AI infrastructure layer", this layer performs preprocessing tasks such as cleaning, handling missing values, normalization, and other necessary transformations, while ensuring data quality and sufficiency for subsequent stages. If required, the data is further annotated with domain-specific labels to enhance its applicability to engineering problems. Ultimately, this layer aims to produce useful and usable data that is well-prepared for downstream processes, enabling efficient and accurate AI model development.

\subsection{Knowledge integration layer}
This layer ensures that domain expertise is integrated into AI models and workflows. This requires a systematic exploration and development of storage, retrieval, sharing, and application of knowledge~\cite{jarrahi2023artificial}. By embedding engineering principles, rules, and domain-specific knowledge into AI model development, this layer enhances the understanding of complex engineering problems and minimizes potential oversights, thereby improving model accuracy, reliability, and relevance to real-world engineering scenarios. Representative techniques include ontology-based modeling, rule-based reasoning systems, knowledge graph construction, and the use of large language models for information retrieval and summarization.

\subsection{Explainability \& trustworthiness layer}
When developing AI models, explainability and trustworthiness are as critical as high model performance in engineering domains. The black-box nature of deep neural networks often raises concerns about how models make predictions, why they make certain decisions, and the reliability of those outcomes. To address these challenges, this layer focuses on two key areas: developing explainable AI (XAI) techniques such as Local Interpretable Model-agnostic Explanations (LIME)~\cite{ribeiro2016should}, and SHapley Additive exPlanations (SHAP) values~\cite{lundberg2017unified}, to clarify model decision-making processes and underlying reasoning~\cite{arrieta2020explainable, ali2023explainable}, and leveraging uncertainty quantification or probabilistic machine learning methods to handle prediction uncertainties and enhance model reliability~\cite{abdar2021review, murphy2022probabilistic}.

\subsection{AI model development layer}
This layer is the heart of the ecosystem framework, supported by "Data foundation layer", "Knowledge integration layer", and "Explainability \& trustworthiness layer", which focus on designing, training, and validating AI models for specific engineering tasks. It involves: (1) selecting suitable AI and ML techniques; (2) developing and optimizing novel algorithms or architectures; (3) training and validating models using engineering benchmarks or real-world data; and (4) managing the AI model lifecycle, including deployment, monitoring, updating, and retraining~\cite{ashmore2021assuring}. A wide range of ML algorithms can be applied in engineering domains, including traditional machine learning algorithms such as tree-based, clustering-based methods and support vector machine, as well as deep learning architectures including convolutional neural networks (CNNs)~\cite{gu2018recent, li2021survey}, recurrent neural networks (RNNs)~\cite{yu2019review}, transformers~\cite{vaswani2017attention, lin2022survey}, and their variants, implemented using platforms like PyTorch and TensorFlow.

\subsection{Evaluation \& feedback layer}
Before AI solutions are implemented in existing engineering systems, the "Evaluation \& feedback layer," as a supplement to the "AI model development layer," focuses on the continuous assessment and refinement of AI solutions~\cite{freeman2020test, Lee2024aunified}. This involves evaluating models using standard performance metrics to identify areas for improvement and, when necessary, developing specialized metrics for specific problems to enhance model accuracy. Additionally, feedback from users and domain experts is collected to further refine AI models. 

\subsection{Deployment \& integration layer}
The last layer is to deploy AI solutions within existing engineering systems, ensuring seamless integration with current workflows and technologies. It also involves monitoring AI model performance post-deployment, as well as refining and optimizing the models in a timely manner to enhance model performance and operational efficiency.~\cite{bosch2021engineering, tatineni2021advanced} The proposed conceptual framework can be applied to diverse engineering fields, including manufacturing, energy, transportation, healthcare, design, computer vision, natural language processing, construction, robotics, mining, agriculture, materials, and bioengineering.

Overall, these layers are interconnected rather than strictly sequential. The process begins with problem identification, along with the foundational support provided by the AI infrastructure layer. At the core of the framework is the AI model development layer, which is directly enabled by data foundation, knowledge integration, and explainability \& trustworthiness layers. These supporting layers enrich the modeling process with high-quality data, domain expertise, and interpretability mechanisms. Following model development, the evaluation \& feedback layer facilitates iterative refinement by monitoring performance and incorporating feedback from both users and real-world deployment results. Finally, the deployment \& integration layer ensures that validated AI models are embedded into real-world engineering systems.

\section{Challenges of Engineering Artificial Intelligence}

Figure~\ref{fig:BasicML} summarizes traditional machine learning methods and algorithms commonly used in engineering domains. These methods have been widely adopted across different fields to address different engineering challenges. In addition, with the advance of deep learning, deep neural network-based approaches are increasingly utilized in diverse engineering applications. Techniques such as CNNs~\cite{gu2018recent, li2021survey}, RNNs~\cite{yu2019review}, transformers~\cite{vaswani2017attention, lin2022survey}, and their variants often demonstrate greater performance when dealing with large amounts of data. However, despite these advancements, the implementation of AI in engineering still faces several challenges, as outlined below:

\begin{figure*}[!ht]
\centering\includegraphics[width=1.0\linewidth]{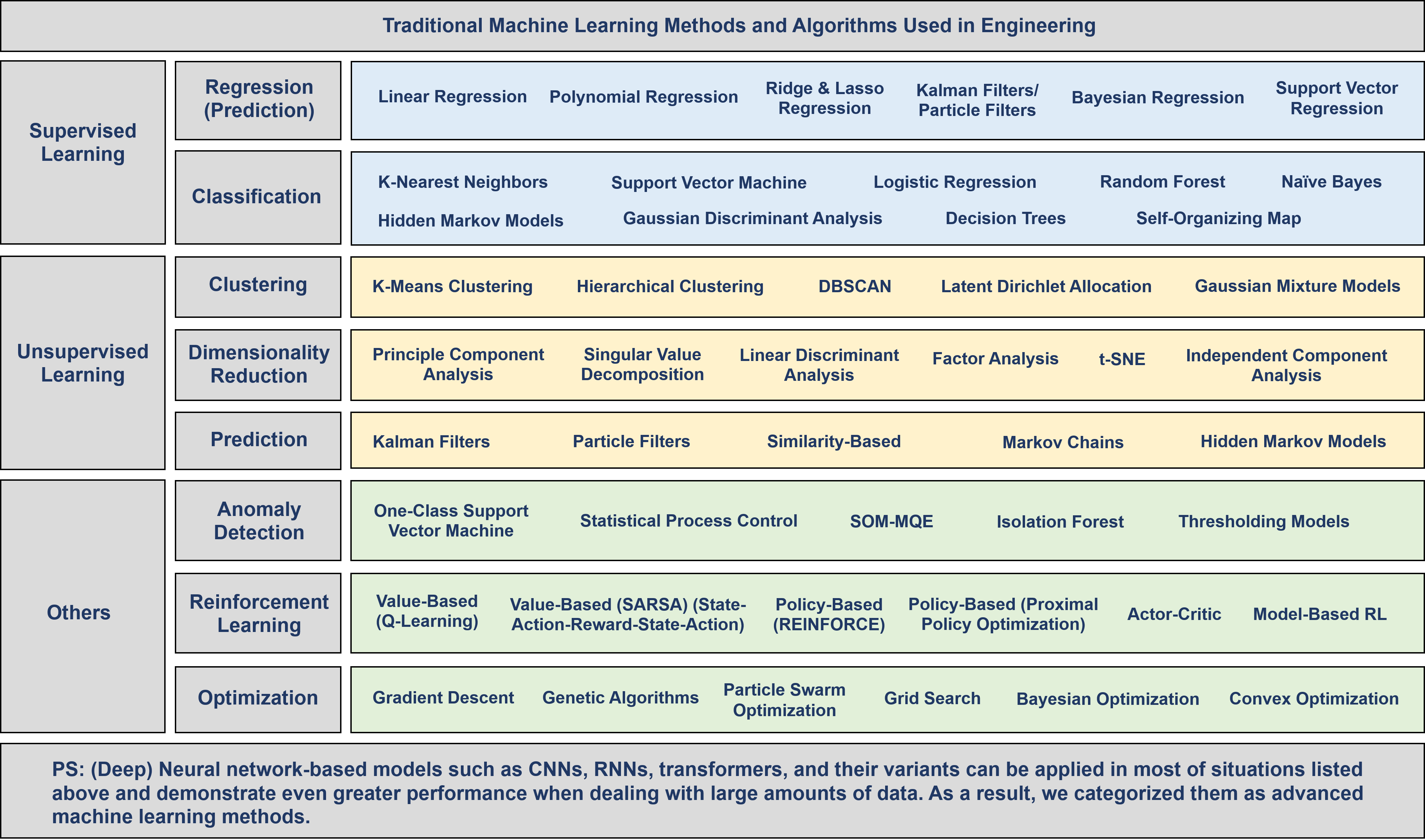}
\caption{Basic Machine Learning Methods and Algorithms Used in Engineering.}
\label{fig:BasicML}
\end{figure*}

\textbf{Data Quality}: The effectiveness of AI models is highly dependent on the quality of the data. Engineering data often suffers from issues such as missing values, noise, inconsistencies, class imbalance, or lack of labels. These challenges necessitate robust preprocessing and validation procedures to ensure the data is useful and useable.

\textbf{Multimodal Data Integration}: Some engineering problems require the integration of data from multiple modalities, such as sensor readings, parametric data, images, or text data. Traditional ML methods often struggle to integrate and process these heterogeneous data types.

\textbf{Domain Shift}: In some engineering applications, data distributions can change due to variations in environmental conditions, operational settings, or other factors. AI models trained on a specific dataset may struggle to maintain accuracy and reliability when faced with new or evolving data distributions.

\textbf{Model Interpretability}: Ensuring that AI models are interpretable remains a significant challenge in engineering, as complex models like deep neural networks are often considered as "black boxes", making it difficult to understand and explain how AI models make predictions or decisions.

\textbf{Model Generalization}: Traditional machine learning models often suffer from overfitting to specific datasets, limiting their ability to perform well on new, unseen data. In engineering AI, achieving robust model performance across diverse scenarios and unseen conditions is essential, requiring the development of models capable of reliable generalization and adaptability.

\begin{figure*}[!ht]
\centering\includegraphics[width=0.8\linewidth]{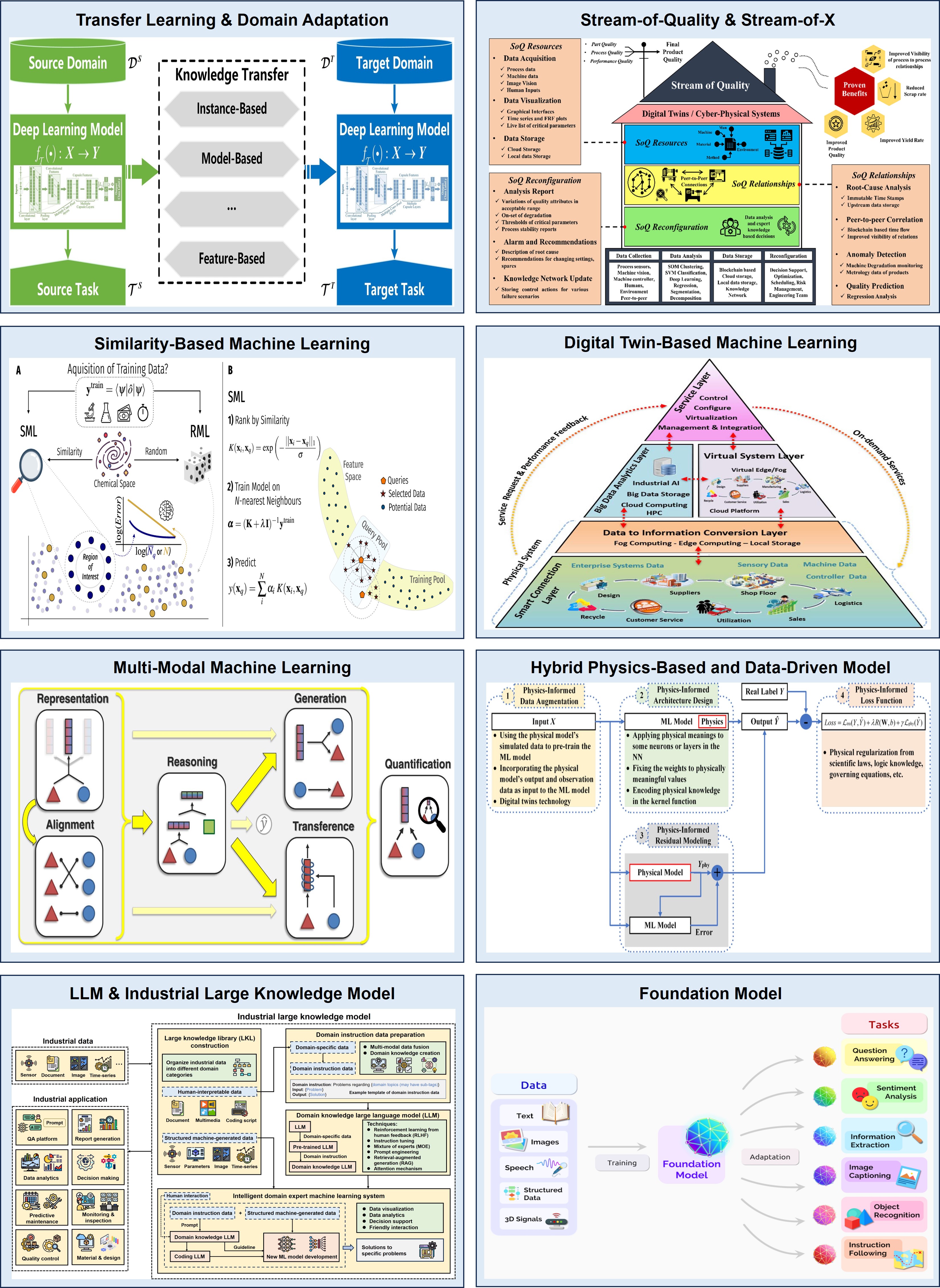}
\caption{Future directions in engineering AI research. Each subfigure is collected from its respective source: (1) Transfer Learning \& Domain Adaptation~\cite{li2022perspective}; (2) Stream-of-Quality \& Stream-of-X~\cite{lee2022stream}; (3) Similarity-Based Machine Learning~\cite{lemm2023improved}; (4) Digital Twin-Based Machine Learning~\cite{lee2020integration}; (5) Multi-Modal Machine Learning~\cite{liang2024foundations}; (6) Hybrid Physics-Based and Data-Driven Models~\cite{li2024review}; (7) Large Language Model \& Industrial Large Knowledge Model~\cite{Lee2024aunified}; (8) Foundation Model~\cite{bommasani2021opportunities}.}
\label{fig:FutureML}
\end{figure*}

\section{Future Direction}
Figure~\ref{fig:FutureML} provides an overview of the emerging future directions in engineering AI. This perspective helps to better understand the key methodologies and advancements and identify critical areas for future research. The eight key research directions are outlined as follows:

\textbf{Transfer Learning \& Domain Adaptation}: enable models trained on one domain to be effectively applied to another, when data in the target domain is scarce or expensive to collect~\cite{farahani2021brief, li2022perspective, ding2023deep}. Future research can focus on refining these techniques to better handle cross-domain applications, limited labeled data in the target domain, and dynamic industrial environments.

\textbf{Stream-of-Quality \& Stream-of-X}: paradigms aim to provide continuous monitoring and optimization of engineering systems for multi-stage manufacturing processes~\cite{lee2022stream, lee2025novel}. Future advancements can integrate real-time analytics with predictive capabilities to enable dynamic quality control, predictive maintenance, and adaptive process management in industrial systems.

\begin{table*}[]
\caption{Qualitative roadmap for future research directions in engineering AI.} \label{table:1}
\begin{center}
\resizebox{\textwidth}{!}{
\renewcommand{\arraystretch}{1.2}
\begin{tabular}{|c|c|c|c|}
\hline
\textbf{Research Direction}                                                            & \textbf{Feasibility} & \textbf{Impact} & \textbf{Leading Sector} \\ \hline
Transfer learning \& domain adaptation & High     & Moderate & Academia + Industry \\ \hline
Similarity-based machine learning      & High     & Moderate & Academia + Industry \\ \hline
Digital twin-based machine learning    & Moderate & High     & Academia + Industry \\ \hline
Stream-of-X                            & Moderate & High     & Academia + Industry \\ \hline
\begin{tabular}[c]{@{}c@{}}Hybrid physics-based and \\ data-driven models\end{tabular} & Low                  & High            & Academia                \\ \hline
Multi-modal machine learning           & Low      & High     & Academia            \\ \hline
LLM and ILKM                           & Low      & High     & Academia + Industry \\ \hline
Foundation models for engineering AI                                                   & Very Low             & Transformative  & Academia + Government   \\ \hline
\end{tabular}}
\end{center}
\end{table*}

\textbf{Similarity-Based Machine Learning}: utilizes similarity measures to identify patterns, relationships, and clusters within data to enhance prediction accuracy~\cite{lemm2023improved, xue2022similarity, niu2022noise}. Future work can explore the way to handle high-dimensional datasets, improve clustering techniques, and integrate explainability features to build trust and reliability in engineering systems.

\textbf{Digital Twin-Based Machine Learning}: By integrating digital twins (combining physical and virtual systems) with ML, this approach enables real-time data analysis, simulation, and prediction~\cite{lee2015cyber,tao2018digital,lee2020integration}. Future developments can focus on developing novel digital twin-based ML to enable real-time monitoring, optimize system performance, and achieve better decision-making through dynamic and accurate system representations.

\textbf{Multi-Modal Machine Learning}: aims to combine information from multiple modalities, such as images, text, audio, tabular data, or sensor signals, to create more robust and comprehensive models~\cite{liang2024foundations, su2023multi, song2024multi}. Future research could explore cross-modal learning efficiency, better representation learning methods, efficient information alignment techniques, etc.

\textbf{Hybrid Physics-Based and Data-Driven Models}: combine the strengths of physics-based modeling and data-driven approaches to improve predictive accuracy, generalizability, and interpretability in complex systems~\cite{wang2022hybrid, li2024review}. Future directions involve developing physics-informed machine learning algorithms, surrogate models, and automating the design and optimization of hybrid models.

\textbf{Large Language Model (LLM) \& Industrial Large Knowledge Model (ILKM)}: LLMs have demonstrated remarkable capabilities in natural language processing, showing the potential for artificial general intelligence (AGI)~\cite{zhao2023survey, chang2024survey}. For future engineering AI, there is a promising opportunity to develop domain-specific ILKMs by integrating structured and unstructured data, knowledge bases, and domain-specific LLMs~\cite{Lee2024aunified}. These models can address complex engineering challenges and facilitate intelligent, data-driven decision-making. 


\textbf{Foundation Model}: is a large-scale ML model, which is pre-trained on massive datasets and can be fine-tuned or adapted for specific downstream tasks with minimal additional training~\cite{bommasani2021opportunities}. The future lies in developing engineering-specific foundation models in different engineering domains such as Prognostics and Health Management (PHM)~\cite{li2024chatgpt}, time series~\cite{goswami2024moment}, CV~\cite{awais2025foundation}, NLP~\cite{touvron2023llama}, healthcare~\cite{moor2023foundation}, manufacturing~\cite{zhang2025large}, etc.

To support strategic planning, Table~\ref{table:1} presents a structured roadmap for future research directions along three dimensions: (1) Feasibility, indicating the maturity and readiness of the technology; (2) Impact, reflecting the expected influence on advancing engineering AI; and (3) Leading sector, including industry, academia, or government. These assessments are based on expert judgment and trends in current literature and practice. Finally, this table serves as an important reference to help researchers and industry practitioners assess maturity, strategic value, investment priority, and collaborative responsibility.

\section{Conclusion}
As AI and ML continue to redefine technological boundaries, there is an urgent need to systematically develop AI solutions in various engineering domains to maximize their impact. This paper identifies the foundational "ABCDE" elements of Engineering AI, proposes a comprehensive conceptual framework for the engineering AI ecosystem, discusses key challenges, and highlights future directions. In conclusion, this work offers a unified and structured pathway for practitioners to implement AI solutions effectively across diverse engineering fields.

\bibliographystyle{abbrv}
\bibliography{manufacturing-letters}

\begin{thebibliography}{10}

\bibitem{abdar2021review}
M.~Abdar, F.~Pourpanah, S.~Hussain, D.~Rezazadegan, L.~Liu, M.~Ghavamzadeh, P.~Fieguth, X.~Cao, A.~Khosravi, U.~R. Acharya, et~al.
\newblock A review of uncertainty quantification in deep learning: Techniques, applications and challenges.
\newblock {\em Information fusion}, 76:243--297, 2021.

\bibitem{ali2023explainable}
S.~Ali, T.~Abuhmed, S.~El-Sappagh, K.~Muhammad, J.~M. Alonso-Moral, R.~Confalonieri, R.~Guidotti, J.~Del~Ser, N.~D{\'\i}az-Rodr{\'\i}guez, and F.~Herrera.
\newblock Explainable artificial intelligence (xai): What we know and what is left to attain trustworthy artificial intelligence.
\newblock {\em Information fusion}, 99:101805, 2023.

\bibitem{arrieta2020explainable}
A.~B. Arrieta, N.~D{\'\i}az-Rodr{\'\i}guez, J.~Del~Ser, A.~Bennetot, S.~Tabik, A.~Barbado, S.~Garc{\'\i}a, S.~Gil-L{\'o}pez, D.~Molina, R.~Benjamins, et~al.
\newblock Explainable artificial intelligence (xai): Concepts, taxonomies, opportunities and challenges toward responsible ai.
\newblock {\em Information fusion}, 58:82--115, 2020.

\bibitem{ashmore2021assuring}
R.~Ashmore, R.~Calinescu, and C.~Paterson.
\newblock Assuring the machine learning lifecycle: Desiderata, methods, and challenges.
\newblock {\em ACM Computing Surveys (CSUR)}, 54(5):1--39, 2021.

\bibitem{awais2025foundation}
M.~Awais, M.~Naseer, S.~Khan, R.~M. Anwer, H.~Cholakkal, M.~Shah, M.-H. Yang, and F.~S. Khan.
\newblock Foundation models defining a new era in vision: a survey and outlook.
\newblock {\em IEEE Transactions on Pattern Analysis and Machine Intelligence}, 2025.

\bibitem{baduge2022artificial}
S.~K. Baduge, S.~Thilakarathna, J.~S. Perera, M.~Arashpour, P.~Sharafi, B.~Teodosio, A.~Shringi, and P.~Mendis.
\newblock Artificial intelligence and smart vision for building and construction 4.0: Machine and deep learning methods and applications.
\newblock {\em Automation in Construction}, 141:104440, 2022.

\bibitem{batra2021emerging}
R.~Batra, L.~Song, and R.~Ramprasad.
\newblock Emerging materials intelligence ecosystems propelled by machine learning.
\newblock {\em Nature Reviews Materials}, 6(8):655--678, 2021.

\bibitem{benos2021machine}
L.~Benos, A.~C. Tagarakis, G.~Dolias, R.~Berruto, D.~Kateris, and D.~Bochtis.
\newblock Machine learning in agriculture: A comprehensive updated review.
\newblock {\em Sensors}, 21(11):3758, 2021.

\bibitem{bommasani2021opportunities}
R.~Bommasani, D.~A. Hudson, E.~Adeli, R.~Altman, S.~Arora, S.~von Arx, M.~S. Bernstein, J.~Bohg, A.~Bosselut, E.~Brunskill, et~al.
\newblock On the opportunities and risks of foundation models.
\newblock {\em arXiv preprint arXiv:2108.07258}, 2021.

\bibitem{bosch2021engineering}
J.~Bosch, H.~H. Olsson, and I.~Crnkovic.
\newblock Engineering ai systems: A research agenda.
\newblock {\em Artificial intelligence paradigms for smart cyber-physical systems}, pages 1--19, 2021.

\bibitem{carvalho2019systematic}
T.~P. Carvalho, F.~A. Soares, R.~Vita, R.~d.~P. Francisco, J.~P. Basto, and S.~G. Alcal{\'a}.
\newblock A systematic literature review of machine learning methods applied to predictive maintenance.
\newblock {\em Computers \& Industrial Engineering}, 137:106024, 2019.

\bibitem{chai2021deep}
J.~Chai, H.~Zeng, A.~Li, and E.~W. Ngai.
\newblock Deep learning in computer vision: A critical review of emerging techniques and application scenarios.
\newblock {\em Machine Learning with Applications}, 6:100134, 2021.

\bibitem{chang2024survey}
Y.~Chang, X.~Wang, J.~Wang, Y.~Wu, L.~Yang, K.~Zhu, H.~Chen, X.~Yi, C.~Wang, Y.~Wang, et~al.
\newblock A survey on evaluation of large language models.
\newblock {\em ACM Transactions on Intelligent Systems and Technology}, 15(3):1--45, 2024.

\bibitem{cheetham2024artificial}
A.~K. Cheetham and R.~Seshadri.
\newblock Artificial intelligence driving materials discovery? perspective on the article: Scaling deep learning for materials discovery.
\newblock {\em Chemistry of Materials}, 36(8):3490--3495, 2024.

\bibitem{ding2023deep}
Y.~Ding, M.~Jia, J.~Zhuang, Y.~Cao, X.~Zhao, and C.-G. Lee.
\newblock Deep imbalanced domain adaptation for transfer learning fault diagnosis of bearings under multiple working conditions.
\newblock {\em Reliability Engineering \& System Safety}, 230:108890, 2023.

\bibitem{dogan2021machine}
A.~Dogan and D.~Birant.
\newblock Machine learning and data mining in manufacturing.
\newblock {\em Expert Systems with Applications}, 166:114060, 2021.

\bibitem{entezari2023artificial}
A.~Entezari, A.~Aslani, R.~Zahedi, and Y.~Noorollahi.
\newblock Artificial intelligence and machine learning in energy systems: A bibliographic perspective.
\newblock {\em Energy Strategy Reviews}, 45:101017, 2023.

\bibitem{farahani2021brief}
A.~Farahani, S.~Voghoei, K.~Rasheed, and H.~R. Arabnia.
\newblock A brief review of domain adaptation.
\newblock {\em Advances in data science and information engineering: proceedings from ICDATA 2020 and IKE 2020}, pages 877--894, 2021.

\bibitem{freeman2020test}
L.~Freeman.
\newblock Test and evaluation for artificial intelligence.
\newblock {\em Insight}, 23(1):27--30, 2020.

\bibitem{goswami2024moment}
M.~Goswami, K.~Szafer, A.~Choudhry, Y.~Cai, S.~Li, and A.~Dubrawski.
\newblock Moment: A family of open time-series foundation models.
\newblock {\em arXiv preprint arXiv:2402.03885}, 2024.

\bibitem{gu2018recent}
J.~Gu, Z.~Wang, J.~Kuen, L.~Ma, A.~Shahroudy, B.~Shuai, T.~Liu, X.~Wang, G.~Wang, J.~Cai, et~al.
\newblock Recent advances in convolutional neural networks.
\newblock {\em Pattern recognition}, 77:354--377, 2018.

\bibitem{guo2022attention}
M.-H. Guo, T.-X. Xu, J.-J. Liu, Z.-N. Liu, P.-T. Jiang, T.-J. Mu, S.-H. Zhang, R.~R. Martin, M.-M. Cheng, and S.-M. Hu.
\newblock Attention mechanisms in computer vision: A survey.
\newblock {\em Computational visual media}, 8(3):331--368, 2022.

\bibitem{jan2023artificial}
Z.~Jan, F.~Ahamed, W.~Mayer, N.~Patel, G.~Grossmann, M.~Stumptner, and A.~Kuusk.
\newblock Artificial intelligence for industry 4.0: Systematic review of applications, challenges, and opportunities.
\newblock {\em Expert Systems with Applications}, 216:119456, 2023.

\bibitem{jarrahi2023artificial}
M.~H. Jarrahi, D.~Askay, A.~Eshraghi, and P.~Smith.
\newblock Artificial intelligence and knowledge management: A partnership between human and ai.
\newblock {\em Business Horizons}, 66(1):87--99, 2023.

\bibitem{khurana2023natural}
D.~Khurana, A.~Koli, K.~Khatter, and S.~Singh.
\newblock Natural language processing: state of the art, current trends and challenges.
\newblock {\em Multimedia tools and applications}, 82(3):3713--3744, 2023.

\bibitem{lee2020integration}
J.~Lee, M.~Azamfar, J.~Singh, and S.~Siahpour.
\newblock Integration of digital twin and deep learning in cyber-physical systems: towards smart manufacturing.
\newblock {\em IET Collaborative Intelligent Manufacturing}, 2(1):34--36, 2020.

\bibitem{lee2015cyber}
J.~Lee, B.~Bagheri, and H.-A. Kao.
\newblock A cyber-physical systems architecture for industry 4.0-based manufacturing systems.
\newblock {\em Manufacturing letters}, 3:18--23, 2015.

\bibitem{lee2022stream}
J.~Lee, P.~Gore, X.~Jia, S.~Siahpour, P.~Kundu, and K.~Sun.
\newblock Stream-of-quality methodology for industrial internet-based manufacturing system.
\newblock {\em Manufacturing Letters}, 34:58--61, 2022.

\bibitem{lee2025novel}
J.~Lee, D.-Y. Ji, and Y.-M. Hsu.
\newblock Novel topological machine learning methodology for stream-of-quality modeling in smart manufacturing.
\newblock {\em Manufacturing Letters}, 2025.

\bibitem{Lee2024aunified}
J.~Lee and H.~Su.
\newblock A unified industrial large knowledge model framework in industry 4.0 and smart manufacturing.
\newblock {\em International Journal of AI for Materials and Design}, 1(2):41--47, 2024.

\bibitem{lee2025rethinking}
J.~Lee and H.~Su.
\newblock Rethinking industrial artificial intelligence: A unified foundation framework.
\newblock {\em International Journal of AI for Materials and Design}, page 025080006, 2025.

\bibitem{lemm2023improved}
D.~Lemm, G.~F. von Rudorff, and O.~A. von Lilienfeld.
\newblock Improved decision making with similarity based machine learning: Applications in chemistry.
\newblock {\em Machine Learning: Science and Technology}, 4(4):045043, 2023.

\bibitem{li2024review}
H.~Li, Z.~Zhang, T.~Li, and X.~Si.
\newblock A review on physics-informed data-driven remaining useful life prediction: Challenges and opportunities.
\newblock {\em Mechanical Systems and Signal Processing}, 209:111120, 2024.

\bibitem{li2022perspective}
W.~Li, R.~Huang, J.~Li, Y.~Liao, Z.~Chen, G.~He, R.~Yan, and K.~Gryllias.
\newblock A perspective survey on deep transfer learning for fault diagnosis in industrial scenarios: Theories, applications and challenges.
\newblock {\em Mechanical Systems and Signal Processing}, 167:108487, 2022.

\bibitem{li2024data}
X.~Li, C.~Yang, C.~M{\o}ller, and J.~Lee.
\newblock Data issues in industrial ai system: A meta-review and research strategy.
\newblock {\em arXiv preprint arXiv:2406.15784}, 2024.

\bibitem{li2024chatgpt}
Y.-F. Li, H.~Wang, and M.~Sun.
\newblock Chatgpt-like large-scale foundation models for prognostics and health management: A survey and roadmaps.
\newblock {\em Reliability Engineering \& System Safety}, 243:109850, 2024.

\bibitem{li2021survey}
Z.~Li, F.~Liu, W.~Yang, S.~Peng, and J.~Zhou.
\newblock A survey of convolutional neural networks: analysis, applications, and prospects.
\newblock {\em IEEE transactions on neural networks and learning systems}, 33(12):6999--7019, 2021.

\bibitem{liang2024foundations}
P.~P. Liang, A.~Zadeh, and L.-P. Morency.
\newblock Foundations \& trends in multimodal machine learning: Principles, challenges, and open questions.
\newblock {\em ACM Computing Surveys}, 56(10):1--42, 2024.

\bibitem{lin2022survey}
T.~Lin, Y.~Wang, X.~Liu, and X.~Qiu.
\newblock A survey of transformers.
\newblock {\em AI open}, 3:111--132, 2022.

\bibitem{lundberg2017unified}
S.~M. Lundberg and S.-I. Lee.
\newblock A unified approach to interpreting model predictions.
\newblock {\em Advances in neural information processing systems}, 30, 2017.

\bibitem{min2023recent}
B.~Min, H.~Ross, E.~Sulem, A.~P.~B. Veyseh, T.~H. Nguyen, O.~Sainz, E.~Agirre, I.~Heintz, and D.~Roth.
\newblock Recent advances in natural language processing via large pre-trained language models: A survey.
\newblock {\em ACM Computing Surveys}, 56(2):1--40, 2023.

\bibitem{moor2023foundation}
M.~Moor, O.~Banerjee, Z.~S.~H. Abad, H.~M. Krumholz, J.~Leskovec, E.~J. Topol, and P.~Rajpurkar.
\newblock Foundation models for generalist medical artificial intelligence.
\newblock {\em Nature}, 616(7956):259--265, 2023.

\bibitem{murphy2022probabilistic}
K.~P. Murphy.
\newblock {\em Probabilistic machine learning: an introduction}.
\newblock MIT press, 2022.

\bibitem{niu2022noise}
C.~Niu, M.~Li, F.~Fan, W.~Wu, X.~Guo, Q.~Lyu, and G.~Wang.
\newblock Noise suppression with similarity-based self-supervised deep learning.
\newblock {\em IEEE transactions on medical imaging}, 42(6):1590--1602, 2022.

\bibitem{nti2022applications}
I.~K. Nti, A.~F. Adekoya, B.~A. Weyori, and O.~Nyarko-Boateng.
\newblock Applications of artificial intelligence in engineering and manufacturing: a systematic review.
\newblock {\em Journal of Intelligent Manufacturing}, 33(6):1581--1601, 2022.

\bibitem{ribeiro2021robotic}
J.~Ribeiro, R.~Lima, T.~Eckhardt, and S.~Paiva.
\newblock Robotic process automation and artificial intelligence in industry 4.0--a literature review.
\newblock {\em Procedia Computer Science}, 181:51--58, 2021.

\bibitem{ribeiro2016should}
M.~T. Ribeiro, S.~Singh, and C.~Guestrin.
\newblock " why should i trust you?" explaining the predictions of any classifier.
\newblock In {\em Proceedings of the 22nd ACM SIGKDD international conference on knowledge discovery and data mining}, pages 1135--1144, 2016.

\bibitem{sahoo2022smart}
S.~Sahoo and C.-Y. Lo.
\newblock Smart manufacturing powered by recent technological advancements: A review.
\newblock {\em Journal of Manufacturing Systems}, 64:236--250, 2022.

\bibitem{sharma2020machine}
A.~Sharma, A.~Jain, P.~Gupta, and V.~Chowdary.
\newblock Machine learning applications for precision agriculture: A comprehensive review.
\newblock {\em IEEE Access}, 9:4843--4873, 2020.

\bibitem{singhal2023large}
K.~Singhal, S.~Azizi, T.~Tu, S.~S. Mahdavi, J.~Wei, H.~W. Chung, N.~Scales, A.~Tanwani, H.~Cole-Lewis, S.~Pfohl, et~al.
\newblock Large language models encode clinical knowledge.
\newblock {\em Nature}, 620(7972):172--180, 2023.

\bibitem{song2024multi}
B.~Song, R.~Zhou, and F.~Ahmed.
\newblock Multi-modal machine learning in engineering design: A review and future directions.
\newblock {\em Journal of Computing and Information Science in Engineering}, 24(1):010801, 2024.

\bibitem{soori2023artificial}
M.~Soori, B.~Arezoo, and R.~Dastres.
\newblock Artificial intelligence, machine learning and deep learning in advanced robotics, a review.
\newblock {\em Cognitive Robotics}, 3:54--70, 2023.

\bibitem{su2024machine}
H.~Su and J.~Lee.
\newblock Machine learning approaches for diagnostics and prognostics of industrial systems using open source data from phm data challenges: A review.
\newblock {\em International Journal of Prognostics and Health Management}, 15(2), 2024.

\bibitem{su2023multi}
H.~Su, B.~Song, and F.~Ahmed.
\newblock Multi-modal machine learning for vehicle rating predictions using image, text, and parametric data.
\newblock In {\em International Design Engineering Technical Conferences and Computers and Information in Engineering Conference}, volume 87295, page V002T02A089. American Society of Mechanical Engineers, 2023.

\bibitem{tao2018digital}
F.~Tao, H.~Zhang, A.~Liu, and A.~Y. Nee.
\newblock Digital twin in industry: State-of-the-art.
\newblock {\em IEEE Transactions on industrial informatics}, 15(4):2405--2415, 2018.

\bibitem{tatineni2021advanced}
S.~Tatineni and A.~Katari.
\newblock Advanced ai-driven techniques for integrating devops and mlops: Enhancing continuous integration, deployment, and monitoring in machine learning projects.
\newblock {\em Journal of Science \& Technology}, 2(2):68--98, 2021.

\bibitem{touvron2023llama}
H.~Touvron, T.~Lavril, G.~Izacard, X.~Martinet, M.-A. Lachaux, T.~Lacroix, B.~Rozi{\`e}re, N.~Goyal, E.~Hambro, F.~Azhar, et~al.
\newblock Llama: Open and efficient foundation language models.
\newblock {\em arXiv preprint arXiv:2302.13971}, 2023.

\bibitem{vaswani2017attention}
A.~Vaswani.
\newblock Attention is all you need.
\newblock {\em Advances in Neural Information Processing Systems}, 2017.

\bibitem{veres2019deep}
M.~Veres and M.~Moussa.
\newblock Deep learning for intelligent transportation systems: A survey of emerging trends.
\newblock {\em IEEE Transactions on Intelligent transportation systems}, 21(8):3152--3168, 2019.

\bibitem{wang2022hybrid}
J.~Wang, Y.~Li, R.~X. Gao, and F.~Zhang.
\newblock Hybrid physics-based and data-driven models for smart manufacturing: Modelling, simulation, and explainability.
\newblock {\em Journal of Manufacturing Systems}, 63:381--391, 2022.

\bibitem{wang2022machine}
X.~Wang, A.~Liu, and S.~Kara.
\newblock Machine learning for engineering design toward smart customization: A systematic review.
\newblock {\em Journal of Manufacturing Systems}, 65:391--405, 2022.

\bibitem{whang2023data}
S.~E. Whang, Y.~Roh, H.~Song, and J.-G. Lee.
\newblock Data collection and quality challenges in deep learning: A data-centric ai perspective.
\newblock {\em The VLDB Journal}, 32(4):791--813, 2023.

\bibitem{xu2021computer}
S.~Xu, J.~Wang, W.~Shou, T.~Ngo, A.-M. Sadick, and X.~Wang.
\newblock Computer vision techniques in construction: a critical review.
\newblock {\em Archives of Computational Methods in Engineering}, 28:3383--3397, 2021.

\bibitem{xue2022similarity}
B.~Xue, H.~Xu, X.~Huang, K.~Zhu, Z.~Xu, and H.~Pei.
\newblock Similarity-based prediction method for machinery remaining useful life: A review.
\newblock {\em The International Journal of Advanced Manufacturing Technology}, 121(3):1501--1531, 2022.

\bibitem{yao2023machine}
Z.~Yao, Y.~Lum, A.~Johnston, L.~M. Mejia-Mendoza, X.~Zhou, Y.~Wen, A.~Aspuru-Guzik, E.~H. Sargent, and Z.~W. Seh.
\newblock Machine learning for a sustainable energy future.
\newblock {\em Nature Reviews Materials}, 8(3):202--215, 2023.

\bibitem{yu2019review}
Y.~Yu, X.~Si, C.~Hu, and J.~Zhang.
\newblock A review of recurrent neural networks: Lstm cells and network architectures.
\newblock {\em Neural computation}, 31(7):1235--1270, 2019.

\bibitem{yuan2022machine}
T.~Yuan, W.~da~Rocha~Neto, C.~E. Rothenberg, K.~Obraczka, C.~Barakat, and T.~Turletti.
\newblock Machine learning for next-generation intelligent transportation systems: A survey.
\newblock {\em Transactions on emerging telecommunications technologies}, 33(4):e4427, 2022.

\bibitem{yuksel2023review}
N.~Y{\"u}ksel, H.~R. B{\"o}rkl{\"u}, H.~K. Sezer, and O.~E. Canyurt.
\newblock Review of artificial intelligence applications in engineering design perspective.
\newblock {\em Engineering Applications of Artificial Intelligence}, 118:105697, 2023.

\bibitem{zhang2022shifting}
A.~Zhang, L.~Xing, J.~Zou, and J.~C. Wu.
\newblock Shifting machine learning for healthcare from development to deployment and from models to data.
\newblock {\em Nature Biomedical Engineering}, 6(12):1330--1345, 2022.

\bibitem{zhang2025large}
H.~Zhang, S.~D. Semujju, Z.~Wang, X.~Lv, K.~Xu, L.~Wu, Y.~Jia, J.~Wu, W.~Liang, R.~Zhuang, et~al.
\newblock Large scale foundation models for intelligent manufacturing applications: a survey.
\newblock {\em Journal of Intelligent Manufacturing}, pages 1--52, 2025.

\bibitem{zhao2023survey}
W.~X. Zhao, K.~Zhou, J.~Li, T.~Tang, X.~Wang, Y.~Hou, Y.~Min, B.~Zhang, J.~Zhang, Z.~Dong, et~al.
\newblock A survey of large language models.
\newblock {\em arXiv preprint arXiv:2303.18223}, 2023.

\end{thebibliography}
\end{document}